\documentclass[twoside,article,norunningheads,natbib,a4paper]{./styles/svproc}

\usepackage{cancel}
\usepackage[normalem]{ulem}
\usepackage{algorithm}
\usepackage{algorithmicx}
\usepackage{algpseudocode} 
\usepackage[cmex10]{amsmath}
\usepackage{mathtools}
\usepackage{amssymb}
\usepackage{balance}
\usepackage{eqparbox}
\usepackage{bbm, flushend}
\usepackage{mathtools}
\usepackage[colorinlistoftodos]{todonotes}
\usepackage{color}
\usepackage{subcaption}
\usepackage{float}
\usepackage{setspace}

\newcommand{\bseq}{\begin{subequations}}
\newcommand{\eseq}{\end{subequations}}
\newcommand{\baln}{\begin{align}}
\newcommand{\ealn}{\end{align}}
\newcommand{\balnd}{\begin{aligned}}
\newcommand{\ealnd}{\end{aligned}}
\newcommand{\beq}{\begin{equation}}
\newcommand{\eeq}{\end{equation}}
\newcommand{\beqn}{\begin{eqnarray}}
\newcommand{\eeqn}{\end{eqnarray}}
\newcommand{\beqno}{\begin{eqnarray*}}
\newcommand{\eeqno}{\end{eqnarray*}}
\newcommand{\bma}{\begin{displaymath}}
\newcommand{\ema}{\end{displaymath}}
\newcommand{\bnu}{\begin{enumerate}}
\newcommand{\enu}{\end{enumerate}}
\newcommand{\bce}{\begin{center}}
\newcommand{\ece}{\end{center}}
\newcommand{\btb}{\begin{tabular}}
\newcommand{\etb}{\end{tabular}}
\newcommand{\ba}{\begin{array}}
\newcommand{\ea}{\end{array}}

\usepackage[numbers]{natbib}
\usepackage{url}

\begin{document}
\mainmatter              
\title{Koopman based trajectory model and computation offloading for high mobility paradigm in ISAC enabled IoT system}
\titlerunning{Koopman trajectory and offloading}  
%
\author{Minh-Tuan Tran}
\authorrunning{Minh-Tuan Tran} 
%
\tocauthor{Minh-Tuan Tran}
\institute{HCMC University of Technology, VNU-HCM, Vietnam
}

\maketitle              

\begin{abstract}
User experience on mobile devices is constrained by limited battery capacity and processing power, but 6G technology advancements are diving rapidly into mobile technical evolution. Mobile edge computing (MEC) offers a solution, offloading computationally intensive tasks to edge cloud servers, reducing battery drain compared to local processing. The upcoming integrated sensing and communication in mobile communication may improve the trajectory prediction and processing delays. This study proposes a greedy resource allocation optimization strategy for multi-user networks to minimize aggregate energy usage. Numerical results show potential improvement at 33\% for every 1000 iteration. Addressing prediction model division and velocity accuracy issues is crucial for better results. A plan for further improvement and achieving objectives is outlined for the upcoming work phase.

\keywords{Koopman, trajectory control, computational offloading, ISAC}
\end{abstract}
\linespread{1.4}
\section{Introduction}
Due to their short battery lives and low processing power, mobile devices and IoT encourage to run computing intensive applications like natural language processing, computer vision, machine learning tasks, (e.g., Apple Siri recognition) on the extra resource \cite{ETSI_2014}. To adapt to changing conditions such as varying workloads, event communication systems combined with application migration \cite{nguyen2012ebc} to enhance system scalability, performance, flexibility, adaptability. The flexibility and adaptive design is widely employed in the managing of complex distributed systems \cite{wan2019task,feng2020dynamic}. As more and more applications require IoT devices to autonomously sense the world around them, the number of these devices will likely increase significantly. In particular, the number of networked devices by the end of 2023 will be 29.3 billion \cite{agarwal2018joint}. Cloud computing has emerged as a solution, offering computational services to these devices. It leads to the need to have edge server placed near the data sources \cite{8016573}, which reduces both response time for computing requests and network load\cite{lyu2016multiuser}. Loosely speaking, the term \emph{edge} is defined as computing resources on the path between these data producers and central clouds. Based on this idea, a system of distributed edge servers can be constructed to support these devices with a proper offloading policy\cite{guo2018energy}. 

However, the burgeoning data generated by these devices is overwhelming network capacities\cite{lin2022energy,nguyen2019computation}.
It is typically necessary to determine a number of offloading system factors before developing such an offloading policy, including channel status information, mobile device parameters, and task information\cite{chen2019toffee}. An ideal optimization algorithm must be fully conscious of the physical characteristics of mobile devices, including processor speed and battery life \cite{nguyen2020joint}. The decision-making mechanism for offloading should be aware of parameters related to the task's computational difficulty; these will be covered in more detail later\cite{bhamare2017optimal}. In addition to the two factors mentioned above, the channel status information has an important role in estimating the task's final offloading ratio\cite{du2017contract}.
Our research works aim to create an effective algorithm to allocate resources for computation offloading techniques on the new multi-access network by combining the classic problem of computation offloading with the advancement in high mobility scenarios.

Given a system consisting a set of $N$ users and one Edge Server (ES). Each user uses exactly one mobile device (MD) then there are total of $N$ MD in the system. On each device, there exists a set of computational tasks. 
These tasks process an input stream of data, and the total number of tasks is denoted by $K$, where each task is indexed by $k \in \{1,2,\dots,K\}$. As a result, input data of a $k$-th task on MD $n$ has the length of $D_{n,k}$ bits.

\vspace{-0.4 cm}\subsection{System notations}\vspace{-0.21 cm}

\noindent
A given system has 1 BS and K mobile devices (MD) connected through wireless network using OTFS modulation which perform the transform velocity $v_k$ and carrier frequency $f_c$ to spectral efficiency in formulation $calcSE(v_k,f_c)$. The BS is equipped an edge server (ES) with unlimited computation capacity. Each MD $k \in K$ has a task with input data size $D_{k}$. At each device, there is a decision of offloading a portion of task to ES z\( l_k D_k\) and the rest \((1-l_k)D_k\) is executed locally on MD.

\vspace{-0.4 cm}\subsection{Computation model}\vspace{-0.21 cm}
\label{sec:Comp_model}
Since it native unlimited capacity, the time and energy consumption in ES to complete offloaded task portion are negligible and can be ignored. The computation is only take in account the portion of task at local MD. We assume MD $n$ operates at a fixed processing speed of \(f_k\)

Each MD is assumed to operate at a fixed processing speed of \(f_n\) during the execution of any task.
To perform computation, the quantity describing the complexity is \(c_{n,k}\) with units of CPU cycles/bit, indicating the average number of clock cycles required to process one bit of input data.
The execution time of the task computed locally on the MD, \( T_{local} \), in seconds, is expressed as follows:

\vspace{-0,3 cm}
\begin{equation}
    \label{eqn:t_local}
    T_{local(n,k)} = \dfrac{c_{n,k} (1 - l_{n,k}) D_{n,k}}{f_{n}}
\vspace{-0.2 cm}
\end{equation}
\noindent
where \(f_{n}\) is the CPU speed of MD \(n\) measured in (Hz), and \(c_{n,k}\) is the number of CPU cycles required to execute one bit of data on MD \(n\) for the task $k$. 

The corresponding amount of energy is:

\vspace{-0,3 cm}
\begin{equation}
    \label{eqn:e_local}
    E_{local(n,k)} = \epsilon_{n} c_{n,k} f_{n}^2 (1 - l_{n,k}) D_{n,k}
\end{equation}
\noindent
where $\epsilon_{n,k}$ is a energy coefficient determined by the chip architecture of MD $n$.

\vspace{-0.4 cm}\subsection{Communication model}\vspace{-0.21 cm}
\label{sec:offloading_model}

Considering that the data size of the task processed result is negligible compared to the data size of the tasks themselves, in this work, we only focus on the uplink transmission where the processed results are usually small. We consider the situation where the ES allocate spectrum bandwidth $W_n$ represents the bandwidth allocated to MD \(n\) by the ES to offload its computation. The uplink transmission rate from MD \(n\) to the edge ES can be rewritten as follows:
\vspace{-0.5 cm}
\begin{equation}
    \label{eqn:comm_rate}
    R_n = W_n \log_2 \left( 1 + \dfrac{p_n h_n}{\sigma^2} \right) = W_n C_{{Zak}}
\vspace{-0.2 cm}
\end{equation}
\noindent
where \(p_n\) is the uplink transmission power of MD \(n\). Assume the paths between transmitter and receiver has little fading effect, the channel gain \(h_n\) is a constant close to 1. \(\sigma^2\) refers to the variance of the additive white Gaussian noise. The notation \(C_{{Zak}} = {calSE}(v_n, f_c)\) (bit/s/Hz) is the spectral efficiency (SE) with variables \(v_n\) being the vehicle speed and \(f_c\) being the carrier frequency.

We assume that the wireless links transmit data at a fixed rate measured in bits per second (bps) with is a common sense \cite{raviteja2019orthogonal}. The transmission delay from MD \(n\) to the ES for offloading tasks can be defined as

\vspace{-0.5 cm}
\begin{equation}
    \label{eqn:t_trans}
    T_{off(n,k)} = \dfrac{l_{n,k} D_{n,k}}{R_{n}} = \dfrac{l_{n,k} D_{n,k}}{W_n {calSE}(v_k, f_c)} 
\vspace{-0.2 cm}
\end{equation}

\noindent
The corresponding energy to perform the transmission is:

\vspace{-0,5 cm}
\begin{equation}
    \label{eqn:e_trans}
    E_{off(n,k)} = p_n T_{off(n,k)} = \left( 2^{{calSE}(v_k, f_c)} - 1\right) \dfrac{\sigma^2}{h_n} \dfrac{l_{n,k} D_{n,k}}{W_n {calSE}(v_k, f_c)}
\vspace{-0.2 cm}
\end{equation}

\vspace{-0.4 cm}\subsection{Problem modeling statement}\vspace{-0.21 cm}
\label{sec:DesignModelProblemStatement}
The estimated total execution time for processing the task $k$ of MD \(n\) for any case can be expressed as follows:

\vspace{-0,8 cm}
\begin{equation}
    \label{eqn:total_time}
    T_{n,k} = T_{off(n,k)} + T_{local(n,k)} =\dfrac{l_{n,k} D_{n,k}}{W_n {calSE}(v_k, f_c)}  + \dfrac{c_{n,k} (1 - l_{n,k}) D_{n,k}}{f_{n}}
\vspace{-0,3 cm}
\end{equation}

The total energy consumption for processing the task of MD \(n\) at time \(k\) is the sum of the component energies
\vspace{-0,3 cm}
\begin{equation} 
\label{formula:e_total}
E_{n,k} = E_{off(n,k)} + E_{local(n,k)}
\vspace{-0,3 cm}
\end{equation}

By substituting the computed quantities from expressions (\ref{eqn:e_local}) and (\ref{eqn:e_trans}) to the equation (\ref{formula:e_total}), we obtain the total energy expression as follows:

\textbf{(Problem P1):}\\
\vspace{-0.7 cm}
\begin{align*}
    \begin{split}
        E_{n,k}&{=} E_{off(n,k)} + E_{local(n,k)} \\
    \          &{=} \left( 2^{{calSE}(v_k, f_c)} - 1\right) \dfrac{\sigma^2}{h_n} \dfrac{l_{n,k} D_{n,k}}{W_n {calSE}(v_k, f_c)} + \epsilon_{n} c_{n,k} f_{n}^2 (1 - l_{n,k}) D_{n,k}
    \end{split}
\vspace{-0.8 cm}
\end{align*}%
\noindent
where the computation of \({calSE}(\cdot)\) is considered an indivisible unit task. This is consistent with the theory that management systems will implement transmission and reception solutions in a modular form designed to operate independently, and system administrators cannot arbitrarily modify the internals of these devices. In other words, they function as black-box components.

\noindent
In the subsequent steps, we will develop the solution for problem \textbf{(P1)}.

\section{Proposed greedy algorithm}
The nature of mobile devices sending offloading requests to the edge server leads to client-server architecture. Mobile devices operate as clients, initiating offloading requests to a dedicated edge server. This interaction occurs over a wireless channel, establishing a bidirectional communication pathway. The server then returns an approval to the request, then the user sends the tasks and related pieces of information to the server and waits for the computation result. 

In the following algorithm, given a list of $N$ users, each with $K$ tasks. The parameters of $k$-th task of $i$-th user is given by:
\begin{itemize}
    \item $d(i,k)$ is the input data length
    \item $c(i,k)$ mobile device computation coefficient J/cycle
    \item $u(i,k)$ task computation coefficient cycle/bit
\end{itemize}

In addition, each mobile user has a fixed computing capacity, i.e. frequency clock rate $f_n$. Initially, the users sends offloading request including its tasks and mobile device's physical parameters. The task is modeled with parameters as above. According to the policy of the optimizing module, after a certain threshold of accepting offloading requests to its task pool. It starts to run the following algorithm. The task pool is implemented as a hash map, in which the key is the mobile device and value is the list of tasks associated with that device.

Initially, it sets the offloading ratio to 0.5 for every tasks. At each step, it recalculates total energy and picks out a task which gives the worst performance, meaning that the energy consumed by this task at the current ratio is not efficient, and then increases the offloading ratio by 10\%. If this increase makes the overall energy consumption by all tasks to decrease, the algorithm continues the next iteration. In the end, the global optimal energy consumption for all tasks is found, but it does not guarantee individual task optimization. The logic for finding task which consumes most energy among all of them is non-trivial, so it is not included here.

There are two primary sources of data generators in this data processing process. The mobile device's trajectory data as well as a set of task-insensitive data. We calculate the estimated channel capacity and communication rate by evaluating the channel condition from the trajectory information. Next, we use the data as the input for the proposed greedy algorithm, which is run on the gateway server, to construct the offloading decision. Finally, we obtain the algorithm's output, which represents the decision to offload.

\linespread{1.12}
\begin{algorithm}[H]
	\caption{Iterative greedy decision} 
         \label{alg:greedyOffload}
\begin{algorithmic}[1]
    \State $N$ is the number of mobile users
    \State $K$ is the number of tasks per user
    \State taskPool: matrix[$K$]
    \State $E_{local}[]$: matrix[$N$]
    \State $E_{offload}[]$: matrix[$N$]
    
    \Function{TotalEnergy}{}
        \State $\text{TotalE} = 0$
        
        \For{md in taskPool}
            \For{task in taskPool[md]}
                \State $taskLen = \text{task.getLength()}$
                \State $mdCompCoef = \text{task.getComp()}$
                \State $taskCompCoef = \text{task.getCoef()}$ 
                \State $E_{\text{local}} = taskLen \times mdCompCoef \times taskCompCoef$
                \State
                \State $B_w =$ this.CommManager.getBandwith()
                \State $h =$ this.CommManager.getGain()
                \State $N_0 =$ this.CommManager.getNoisePower()
                \State $SE =$ this.CommManager.getSpectralEfficiency()
                \State $t = taskLen / (SE * B_w)$
                \State $\text{TotalE} = \text{TotalE} + E_{\text{local}} + E_{\text{offload}}$
            \EndFor
        \EndFor
        
        \State \textbf{return} TotalE
    \EndFunction
    
    \Function{optimize}{}
        \For{md in taskPool}
            \For{task in taskPool[md]}
                \State \text{task.setOffloadingRatio}(0.5)
            \EndFor
        \EndFor
        
        \State $\text{currE} = \text{newE} = \text{this.TotalEnergy()}$
        
        \While{\text{True}}
            \State \text{worst\_task} = \text{this.find\_worst()}
            \State \text{worst\_task.increaseOffload}(0.1)
            
            \State $\text{currE} = \min(\text{currE}, \text{newE})$
            \State $\text{newE} = \text{this.TotalEnergy()}$
            
            \If{\text{newE} $\geq$ \text{currE}}
                \State \text{break}
            \EndIf
        \EndWhile
        
        \State \textbf{return}
    \EndFunction
\end{algorithmic}

\end{algorithm}

\section{Numerical Results}
\subsection{VED Dataset}
We consider a novel large-scale database of the mobile device in real-world. We use the VED dataset \cite{ved}, which records GPS trajectories of automobiles along with historical data on fuel, energy, speed, and auxiliary power consumption, since our prediction primarily takes into account mobile devices traveling in longitude and lattice coordination. Although, the mobile device feature the 3D trajectory, we have exam the UAV quadrator in our practice setting and figure out that the energy consumption is extremely large in tradeoff the 3D moving. Then, we are interested in the model of UAV in plane shape where it moving in 2D and stablize the height in a long duration task. Therefore,the longitude/latitude data in the dataset is fit for our experiment input.
The dataset can be accessed at \url{https://github.com/gsoh/VED}.

\begin{figure}[ht]
    \centering
    \includegraphics[width=0.90\linewidth]{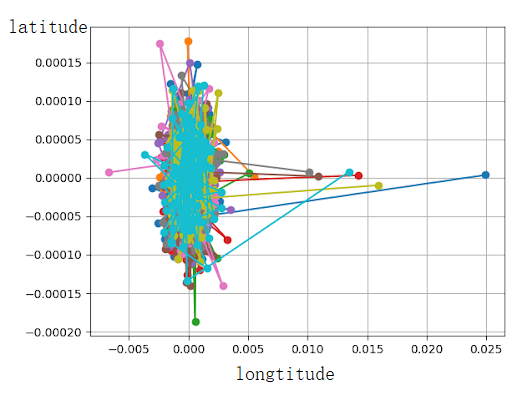}
    \caption{Longtidue and latitude variant in VED dataset}
    \label{fig:trajDatasetPlot}
\end{figure}

Although the dataset accumulates approximately 374,000 miles long, we re-estimate the distance between 2 timestamp which is pre-set at 1000 to 4000 ms. Using the GPS transformation, we conduct an analysis on the gps location of mobile device in which current states are defined as latitude or longitude of as shown in Figure \ref{fig:trajDatasetPlot}. 

\begin{figure}[ht]
    \centering
    \includegraphics[width=0.55\linewidth]{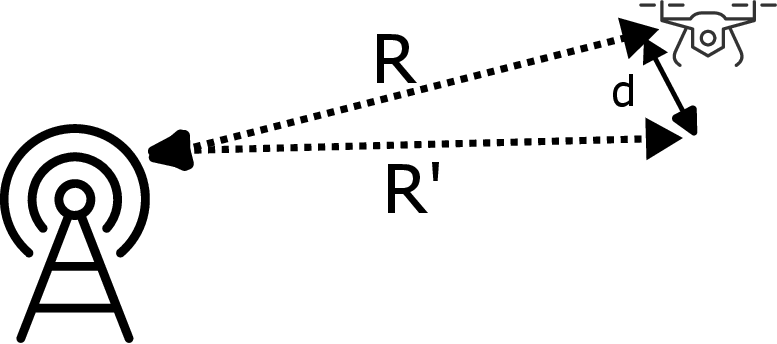}
    \caption{The moving distant vs signal radius distant}
    \label{fig:traRep}
\end{figure}

We also calculate that the greatest distance between two points, or the trajectory between the two sibling timestamps, is at most 400 meters, and that latitude varies more than longitude. Given that the traveling distance is measured in meters and the base station's radius ranges from 3 to 6.5 km, or around 12 km in the both sides. As in figure \ref{fig:traRep}, we observe the distant d is significantly lower than the radius R, we can confirm that R is equivalent to R' as our assumption in prediction model. In summary, we confirm the assuption of the moving is 1 dimension variable of the anglar $\phi$

\subsection{The convergence of Algorithm \ref{alg:greedyOffload} and the impact of offloading strategies}

\begin{figure}[h]
    \centering
    \includegraphics[width=0.65\linewidth]{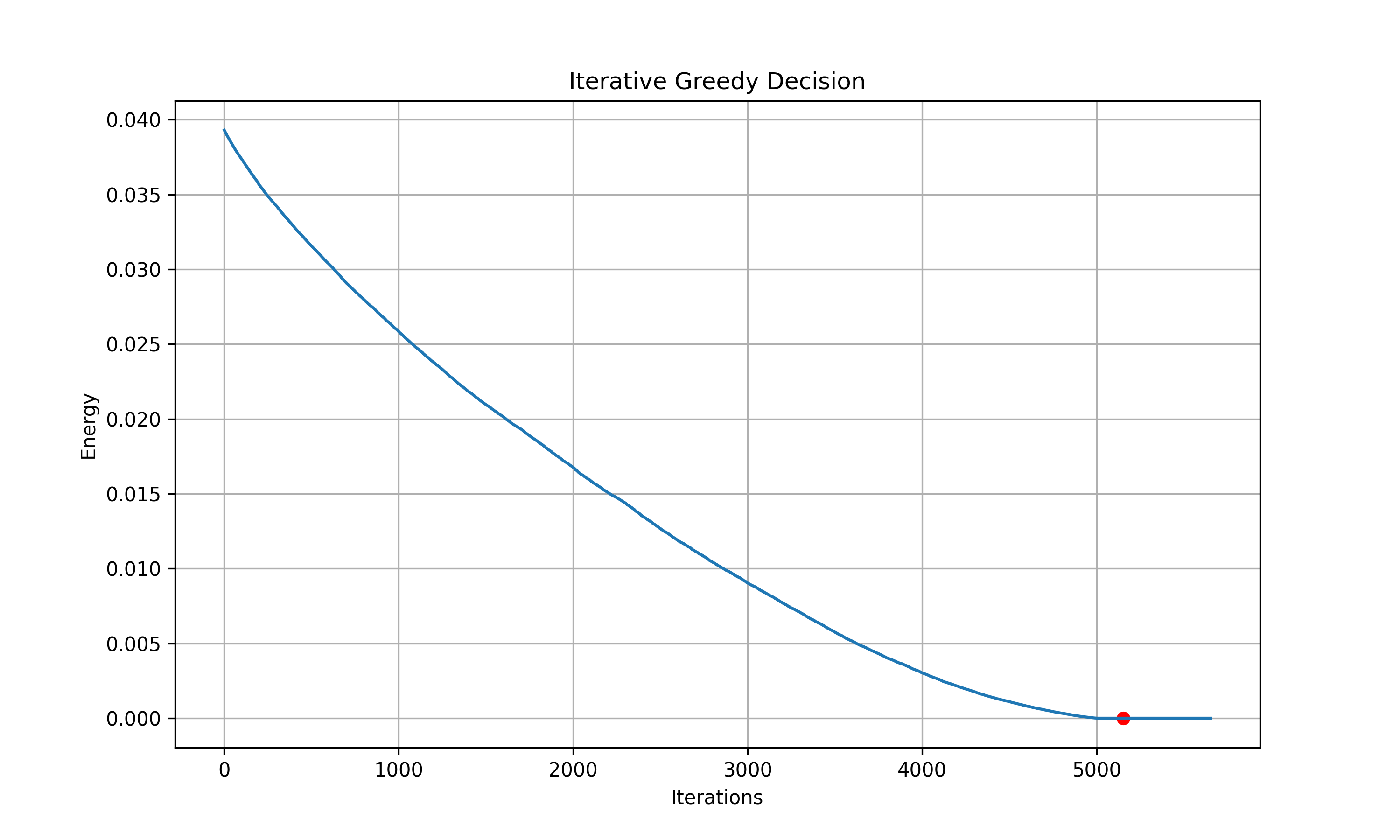}
    \vspace{-0.2cm}
    \caption{Greedy energy consumption optimization algorithm}
    \vspace{-0.2cm}
    \label{fig:GreedyConvergence}
\end{figure}

By choosing locally optimal energy and using an iterative approach to an optimal solution, the greedy algorithm aims to reduce the total energy. The greedy offloading approach, as illustrated in Figure \ref{fig:GreedyConvergence}, takes decisions based on immediate reduction, which definitely leads to a convergence finalization. Furthermore, its ignorance of the whole picture result in unconfined converge to the globally optimal solution of the greedy heuristic; instead, the algorithm may become stuck at a sub-optimal answer. Because of the un-guaranteed step choices, we experience an unpredictable decreasing step of approximately 30\% after every 1000 iteration. Neither the convergence rate nor the bound are established.

Since it is crucial to the edge data-driven network, the impact of data size on energy consumption needs to be considered in the next assessment step. The rapid rise in data size results in a considerable increase in the energy necessary for data transmission. 
Since mobile edge computing usually prefers to outsource tasks to the server, it is essential to figure out how to consume less energy while yet delivering the best results in a limit amount of time. In our initial study phase, we examined the first energy factor; the remaining factors were left for future research development in the subsequent phase.
We conducted an experiment where a higher demand increased the requirement for energy offloading and computation, leading to a larger decrease in the amount of energy used in our optimal strategy.
The better energy saving gap in increasing the offloading data portion is confirmed by the improved gap (the larger gap is the better).

\subsection{Communication rate estimation in velocity modulation}
\begin{figure}[H]
    \centering
    \includegraphics[width=0.7\linewidth]{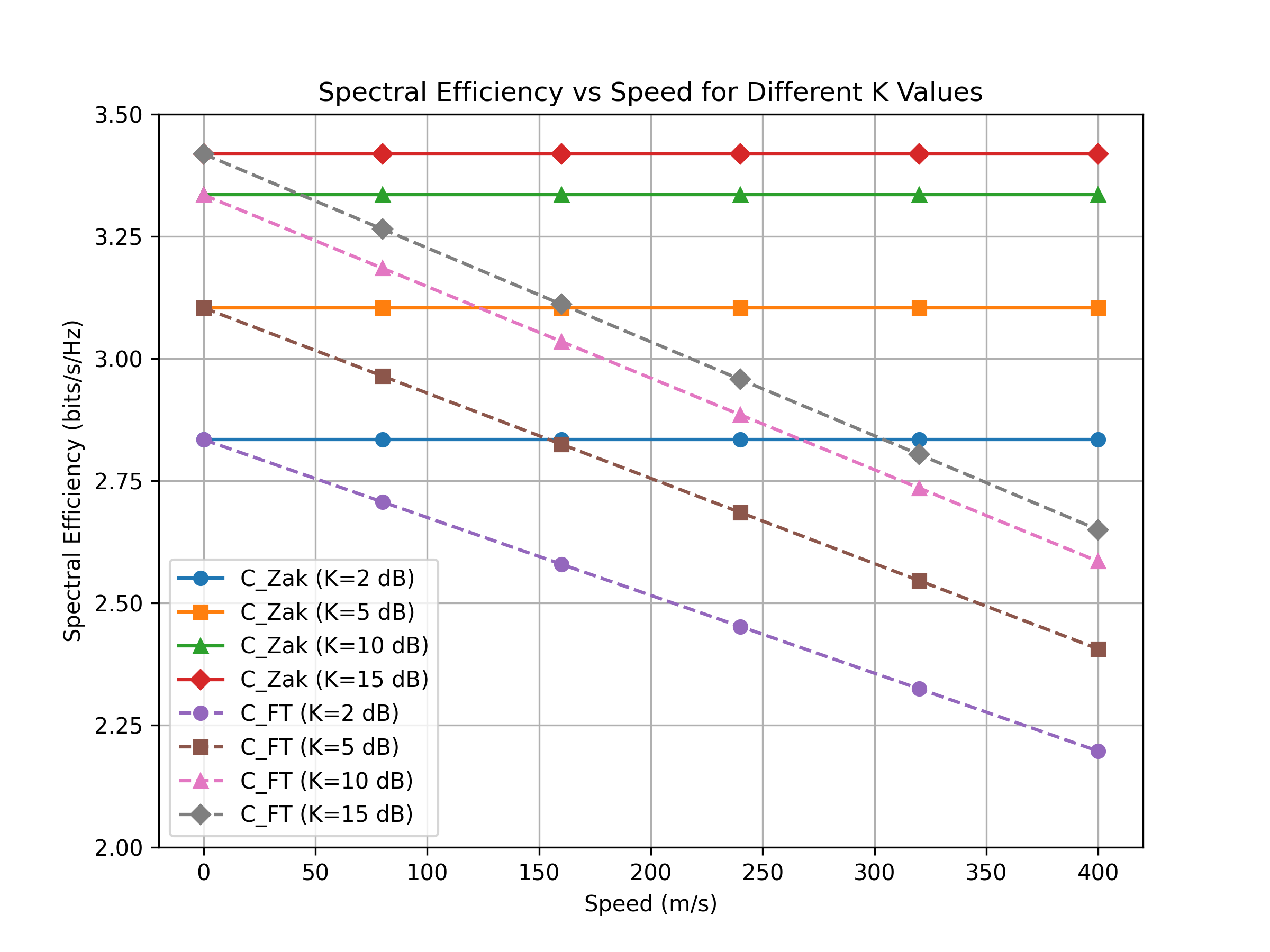}
    \caption{Spectral efficieny of Zak-based and SFFT-based modulation}
    \label{fig:exp-spectral}
\end{figure}

We would want to clarify that that we only leverage the research results of velocity modulation from other group in our research team.Consequently, we don't conduct any additional research and instead  use the provided estimation as a parameter of our work. We want to examine
the spectral efficiency (SE) of the two-step receiver and the result implies the decreasing of communication rate as the speed increases as in Figure \ref{fig:exp-spectral}. We figure out that the SE is calculated in bit/s/Hz which mean in each different settings of the allocated bandwith (in Hz) we achieve the different communication rate.

\subsection{The estimation of offloading energy efficiency}
\begin{figure}[h]
\centering
\includegraphics[width=0.7\textwidth]{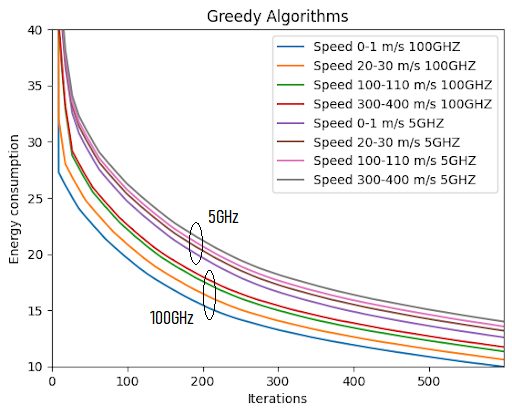}
\vspace{-0.2 cm}
\caption{Greedy energy consumption optimization algorithm in various settings of velocity and bandwidth} 
\vspace{-0.2cm}
\label{fig:SpeedBWEnergyConsumption}
\end{figure}

In the previous evaluation, the system's efficiency is measured by the total energy consumption of mobile devices and processing of their tasks. We only consider the amount of data need to be uploaded has an important impact on the overall system performance. Since the communication settings are also needed to evaluation, we derive the following experiment to figure out the velocity impact which result in the communication rate and even the bandwidth we allocate for the offloading communication.
In Figure. \ref{fig:SpeedBWEnergyConsumption}, the system's efficiency is measured by the total energy consumption of mobile devices and the different settings of velocity and bandwidth.

We figure out that a reduced energy consumption results from the mobile device's high speed, which offers a bigger capacity of communication rate. This can be explained by the increased speed; velocity modulation shows a higher rate of communication, which when combined with the previous offloading method finding, encourages more offloading in order to reduce energy usage. Therefore, we get the lower energy consumption as shown in the illustration. 

\section{Conclusion}
\vspace{-0.4cm}
The evaluation show the impact of offloading decision on the total energy consumption. The impact are further investigated to see the transmission rate caused by velocity estimation. To give the overal view of estimation, we come with trajectory and velocity modeling experiment

The evaluation shows how the offloading strategies affects the total energy consumption.  The impact are further investigated by determining the transmission rate which is induced by velocity modulation. We aims to conduct a model that helps predict the velocity in order to automatic and timely estimate the communication rate. Unfortunately, we obtain the result of low accuracy in our model, we initially figure out that we got some wrong in implementation the Koopman model, that might effect the final result. We have to make a checkpoint the status in this specialized project to 

The analysis illustrates the impact of the offloading strategies on the total energy consumption.  By figuring out the transmission rate that is prompted by velocity modulation, its impacts are further investigated. Our goal is to develop a model which helps in velocity predictions so that the communication rate may be automatically and promptly estimated. Unfortunately, we discover that \textbf{our model's accuracy is low}. Initially, we suspect that there were some implementation errors with the Koopman model, which could have an impact on the final results. To keep the finalize of this specialized project, we have to make a checkpoint of the current status let it a further investigation later.

We still lack of proposing a solution that combines the two sub-problems. We have no strategy to determine the starting point that would enable us to guarantee a suitable and sufficient beginning point. Lastly, the most crucial step in our decision-making process—establishing a constrained statement of convergence—is currently missing.

\bibliographystyle{./styles/bibtex/splncs03_unsrt}
\renewcommand{\bibname}{References}
\bibliography{ms}
\end{document}